\documentclass[11pt,letterpaper]{article}
\usepackage{emnlp2016}
\usepackage{times}
\usepackage{latexsym}
\usepackage{amsmath}
\usepackage{amssymb}
\usepackage{graphicx}
\usepackage{array}
\newcolumntype{C}[1]{>{\centering\let\newline\\\arraybackslash\hspace{0pt}}m{#1}}
\emnlpfinalcopy

\def\emnlppaperid{***}

\title{Structured prediction models for RNN based sequence labeling in clinical text}
\author{Abhyuday N Jagannatha$^1$, Hong Yu$^{1,2}$\\
$^1$ University of Massachusetts, MA, USA\\
$^2$ Bedford VAMC and CHOIR, MA, USA\\
{\tt abhyuday@cs.umass.edu} , {\tt hong.yu@umassmed.edu}\\
}

\date{}

\begin{document}

\maketitle

\begin{abstract}
Sequence labeling is a widely used method for named entity recognition and information extraction from unstructured natural language data. In clinical domain one major application of sequence labeling involves extraction of medical entities such as medication, indication, and side-effects from Electronic Health Record narratives. Sequence labeling in this domain, presents its own set of challenges and objectives. In this work we experimented with various CRF based structured learning models with Recurrent Neural Networks. We extend the previously studied LSTM-CRF models with explicit modeling of pairwise potentials. We also propose an approximate version of skip-chain CRF inference with RNN potentials. We use these methodologies\footnote{Code will be available soon at https://github.com/abhyudaynj/LSTM-CRF-models} for structured prediction in order to improve the exact phrase detection of various medical entities.
\end{abstract}

\section{Introduction}
Patient data collected by hospitals falls into two categories, structured data and unstructured natural language texts. It has been shown that natural text medical data such as discharge summaries, progress notes, etc are rich sources of medically relevant information like adverse drug events, medication prescriptions, diagnosis information etc. Information extracted from these natural text documents can be useful for a multitude of purposes ranging from drug efficacy analysis to adverse effect surveillance. 

A widely used method for Information Extraction from natural text documents involves treating the text as a sequence of tokens. This format allows various sequence labeling algorithms to label the relevant information that should be extracted. Several sequence labeling algorithms such as Conditional Random Fields (CRFs), Hidden Markov Models (HMMs), Neural Networks have been used for information extraction from unstructured text. CRFs and HMMs are probabilistic graphical models that have a rich history of Natural Language Processing (NLP) related applications. These methods try to jointly infer the most likely label sequence for a given sentence.

Recently Recurrent (RNN) or Convolutional Neural Network (CNN) models have increasingly been used for various NLP related tasks. These Neural Networks by themselves, however, do not treat sequence labeling as a structured prediction problem. Different Neural Network models use different methods to synthesize a context vector for each word. This context vector contains information about the current word and its neighboring content. In the case of CNN, the neighbors comprise of words in the same filter size window, while in Bidirectional-RNNs (Bi-RNN) they contain the entire sentence. 

Graphical models and Neural Networks have their own strengths and weaknesses. While graphical models predict the entire label sequence jointly, they usually require special hand crafted features to provide good results. Neural Networks (especially Recurrent Neural Networks), on the other hand, have been shown to be extremely good at identifying patterns from noisy text data, but they still predict each word label in isolation and not as a part of a sequence. In simpler terms, RNN benefit from recognizing patterns in the surrounding input features, while structured learning models like CRF benefit from the knowledge about neighboring label predictions. Recent works on Named Entity Recognition by  \cite{huang2015bidirectional} and others have combined the benefits of Neural Networks with CRF by modeling the unary potential functions of a CRF as NN models. They model the pairwise potentials as a Matrix $[A]$ where the entry $A_{i,j}$ corresponds to the transition probability from the label $i$ to label $j$. Incorporating CRF inference in Neural Network models helps in labeling exact boundaries of various named entities by enforcing pairwise constraints. 

This work focuses on labeling medical events (medication, indication, and adverse drug events) and event related attributes (medication dosage, route, etc) in unstructured clinical notes from Electronic Health Records. Later on in the Section 4, we explicitly define the medical events and attributes that we evaluate on. In the interest of brevity, for the rest of the paper, we use the broad term ``Medical Entities" to refer to all medically relevant information that we are interested in labeling. 

Detecting medical entities in medical documents such as Electronic Health Record notes composed by clinicians presents a somewhat different set of challenges than similar sequence labeling applications in NLP such as Named Entity Recognition. This difference is partly due to the critical nature of medical domain, and partly due to the nature of medical texts and entities therein. Firstly, in the medical domain, extraction of exact medical phrase is extremely important. The names of medical entities often follow polynomial nomenclature. Disease names such as \textit{Uveal melanoma} or \textit{hairy cell leukemia} need to be identified exactly, since partial names ( \textit{hairy cell} or \textit{melanoma}) might have significantly different meanings. Additionally, important medical entities can be relatively rare events in Electronic Health Records. For example, mentions of Adverse Drug Events occur once every six hundred words in our corpus. CRFs inference with NN models cited previously do improve exact phrase labeling. However, better ways of modeling the pairwise potential functions of CRFs might led to improvements in labeling rare entities and detecting exact phrase boundaries. 

Another important challenge in this domain is a need to model long term label dependencies.  For example, in the sentence \textit{``the patient exhibited \textbf{\textit{A}} secondary to \textbf{\textit{B}}"}, the label for \textbf{\textit{A}} is strongly related to the label prediction of \textbf{\textit{B}}. \textbf{\textit{A}} can either be labeled as an adverse drug reaction or a symptom if \textbf{\textit{B}} is a Medication or an Diagnosis respectively. Traditional linear chain CRF approaches that only enforce local pairwise constraints might not be apt to model these dependencies. It can be argued that RNNs may implicitly model label dependencies through patterns in input features of neighboring words. While this is true, explicitly modeling the long term label dependencies can be expected to perform better.

In this work, we explore various methods of structured learning using RNN based feature extractors. We use LSTM as our RNN model. Specifically, we model the CRF pairwise potentials using Neural Networks. We also model an approximate version of skip chain CRF to capture the aforementioned long term label dependencies. We show that these modified frameworks improve the performance when compared to standard LSTM or CRF-LSTM models with the same number of trainable parameters. To the best of our knowledge, this is the only work focused on usage and analysis of RNN based structured learning techniques on extraction of medical entities from clinical notes. 

\section{Related Work}
As mentioned in the previous sections, both Neural Networks and Conditional Random Fields have been widely used for sequence labeling tasks in NLP. Specially, CRFs \cite{lafferty2001conditional} have a long history of being used for various sequence labeling tasks in general and named entity recognition in particular. Some early notable works include McCallum et. al. \shortcite{mccallum2003early}, Sarawagi et al. \shortcite{sarawagi2004semi} and Sha et. al. \shortcite{sha2003shallow}. Hammerton et. al. \shortcite{hammerton2003named} and Chiu et. al. \shortcite{chiu2015named} used Long Short Term Memory (LSTM) \cite{hochreiter1997long} for named entity recognition.

Several recent works on both image and text based domains have used structured inference to improve the performance of Neural Network based models. In NLP, Collobert et al \shortcite{collobert2011natural} used Convolutional Neural Networks to model the unary potentials. Specifically for Recurrent Neural Networks,  Lample et al. \shortcite{lample2016neural} and Huang et. al. \shortcite{huang2015bidirectional} used LSTMs to model the unary potentials of a CRF. 

In biomedial named entity recognition, several approaches use a biological corpus annotated with entities such as protein or gene name. Settles \shortcite{settles2004biomedical} used Conditional Random Fields to extract occurrences of protein, DNA and similar biological entity classes. Li et. al. \shortcite{li2015biomedical} recently used LSTM for named entity recognition or protein/gene names from BioCreative corpus. Gurulingappa et. al. \shortcite{gurulingappa2010empirical} evaluated various existing biomedical dictionaries on extraction of adverse effects and diseases from a corpus of Medline abstracts. 

Our work uses a real world clinical corpus of Electronic Health Records annotated with various medical entities. Other works using a real world medical corpus include Rochefort et al. \shortcite{rochefort2015novel}, who worked on narrative radiology reports. They used a SVM-based classifier with bag of words feature vector to predict deep vein thrombosis and pulmonary embolism. Miotto et. al. \shortcite{miotto2016deep} used a denoising autoencoder to build an unsupervised representation of Electronic Health Records which could be used for predictive modeling of patient's health.


\section{Methods}
We use Bi-RNNs as the feature extractors from the word sequence. We evaluate three different methodologies of structured learning. The baseline is a bidirectional recurrent neural network as described in Section 3.1. 
\subsection{Bi-LSTM (baseline)} 
This model is a standard bidirectional LSTM Neural Network with Word Embedding Input and a Softmax Output layer. The raw natural language input sentence is processed with a regular expression tokenizer into sequence of tokens $\textbf{x}=[x_t]_1^T$. The token sequence is fed into the embedding layer, which produces dense vector representation of words. The word vectors are then fed into a bidirectional RNN layer. This bidirectional RNN along with the embedding layer is the main machinery responsible for learning a good feature representation of the data. The output of the bidirectional RNN produces a feature vector sequence $\boldsymbol{\omega}(\textbf{x})= [\omega(\textbf{x})]_1^T$ with the same length as the input sequence $\textbf{x}$. In this baseline model, we do not use any structured inference. Therefore this model alone can be used to predict the label sequence, by scaling and normalizing $[\omega(\textbf{x})]_1^T$. This is done by using a softmax output layer, which scales the output for a label $l$ where $l \in \{1,2, ...,L\}$ as follows: 
\begin{equation}
P(\tilde{y}_t=j|\textbf{x})=\frac{\exp(\omega(\textbf{x})_t W_j)}{\sum_{l=1}^L \exp(\omega(\textbf{x})_t W_l)}
\end{equation}
The entire model is trained end to end using categorical cross-entropy loss. 

\subsection{Bi-LSTM CRF}
This model is adapted from the Bi-LSTM CRF model described in Huang et. al. \shortcite{huang2015bidirectional}. It combines the framework of bidirectional RNN layer$[\omega(\textbf{x})]_1^T$  described above, with linear chain CRF inference. For a general linear chain CRF the probability of a label sequence $\tilde{y}$ for a given sentence $\textbf{x}$ can be written as  :
\begin{equation}
P(\tilde{y}|\textbf{x})=\frac{1}{Z}\prod_{t=1}^{N}\exp\{\phi(\tilde{y}_{t}) + \psi(\tilde{y}_t,\tilde{y}_{t+1}) \}
\end{equation}
Where $\phi({y}_{t})$ is the unary potential for the label position $t$ and $\psi({y}_t,{y}_{t+1})$ is the pairwise potential between the positions \textit{t,t+1}. Similar to Huang et. al. \shortcite{huang2015bidirectional}, the outputs of the bidirectional RNN layer $\boldsymbol{\omega}(\textbf{x})$ are used to model the unary potentials of a linear chain CRF.  In particular, the NN based unary potential $\phi_{nn}({y}_{t})$ is obtained by passing $\boldsymbol{\omega}(\textbf{x})_{t}$ through a standard feed-forward $tanh$ layer. The binary potentials or transition scores are modeled as a matrix $[A]_{L \times L}$. Here $L$ equals the number of possible labels including the \textit{Outside} label. Each element $A_{i,j}$ represents the transition score from label $i$ to $j$. The probability for a given sequence $\tilde{y}$ can then be calculated as :
\begin{equation}
P(\tilde{y}|\textbf{x};\theta)=\frac{1}{Z}\prod_{t=1}^{T}\exp\{ \phi_{nn}(\tilde{y}_{t}) + A_{\tilde{y_t},\tilde{y}_{t+1}}\}
\end{equation}

The network is trained end-to-end by minimizing the negative log-likelihood of the ground truth label sequence $\hat{\textbf{y}}$ for a sentence $\textbf{x}$ as follows:
\begin{equation}
\mathcal{L}(\textbf{x},\hat{\textbf{y}};\theta)=-\sum_t \sum_{y_t} \delta(y_t=\hat{y}_t) \log{P(y_t|\textbf{x};\theta)}\}
\end{equation}

\subsection{ Bi-LSTM CRF with pairwise modeling}
In the previous section, the pairwise potential is calculated through a transition probability matrix $[A]$ irrespective of the current context or word. For reasons mentioned in section 1, this might not be an effective strategy. Some Medical entities are relatively rare. Therefore transition from an \textit{Outside} label to a medical label might not be effectively modeled by a fixed parameter matrix. In this method, the pairwise potentials are modeled through a non-linear Neural Network which is dependent on the current word and context. Specifically, the pairwise potential $\psi({y}_t,y_{t+1})$ in equation 2 is computed by using a one dimensional CNN with 1-D filter size 2 and $tanh$ non-linearity. At every label position $t$, it takes $[ \omega(\textbf{x})_t ; \omega(\textbf{x})_{t+1} ]$ as input and produces a $L \times L$ pairwise potential output $\psi_{nn}({y}_t,y_{t+1})$

The unary potential calculation is kept the same as in Bi-LSTM-CRF. Substituting the neural network based pairwise potential $\psi_{nn}(\textbf{x})t_{t,t+1}$ into equation 2 we can reformulate the probability of the label sequence $\tilde{y}$ given the word sequence $\textbf{x}$ as : 
\begin{equation}
P(\tilde{y}|\textbf{x};\theta)=\frac{1}{Z}\prod_{t=1}^{N}\exp\{\phi_{nn}(\tilde{y}_{t}) + \psi_{nn}(\tilde{y}_t,\tilde{y}_{t+1}) \}
\end{equation} 

The neural network is trained end-to-end with the objective of minimizing the negative log likelihood in equation 4. 

\begin{table}
\small
\centering
\begin{tabular}{| l | C{1.7cm} | C{2.5cm} |}
\hline \bf Labels & \bf Num. of Instances & \bf Avg word length $\pm$ std  \\ \hline \hline
ADE&1807&1.68 $\pm$ 1.22 \\ \hline
Indication&3724&2.20 $\pm$ 1.79 \\ \hline
Other SSD&40984&2.12 $\pm$ 1.88 \\ \hline
Severity &3628 &1.27 $\pm$ 0.62 \\ \hline
Drugname&17008&1.21 $\pm$ 0.60 \\ \hline
Duration&926&2.01 $\pm$ 0.74 \\ \hline
Dosage&5978&2.09 $\pm$ 0.82 \\ \hline
Route&2862&1.20$\pm$ 0.47 \\ \hline
Frequency&5050&2.44$\pm$ 1.70 \\ \hline
\end{tabular}
\caption{\label{data-table} Annotation statistics for the corpus.}
\end{table}

\subsection{Approximate Skip-chain CRF}

Skip chain models are modifications to linear chain CRFs that allow long term label dependencies through the use of \textit{skip edges}. These are basically edges between label positions that are not adjacent to each other. Due to these skip edges, the skip chain CRF model \cite{sutton2006introduction} explicitly models dependencies between labels which  might be more than one positions apart. The joint inference over these dependencies are taken into account while decoding the best label sequence. However, the loopy graph in  skip chain CRF renders exact inference intractable. Approximate solutions to inference in such models require multiple iterations of loopy belief propagation (BP). Since, each gradient descent iteration for a combined RNN-CRF model requires a fresh calculation of the marginals, this approach is very computationally expensive.  In one approach to mitigate this, Lin et. al. \shortcite{lin2015deeply} directly model the messages in the message passing inference of a 2-D grid CRF for image segmentation. This bypasses the need for modeling the potential function, as well as calculating the approximate messages on the graph using loopy BP. 

\begin{table*}
\small
\centering
\begin{tabular}{| *{7}{ c |}}
\cline{2-7} 
\multicolumn{1}{c}{} & \multicolumn{3}{| c }{\textbf{Strict Evaluation ( Exact Match)}} & \multicolumn{3}{| c |}{\textbf{Relaxed Evaluation (Word based)}} \\ \cline{2-7} \hline
\textbf{Models} & \textbf{Recall} & \textbf{Precision} & \textbf{F-score} & \textbf{Recall} & \textbf{Precision} & \textbf{F-score} \\ \hline
Bi-LSTM &0.8101&0.7845&0.7971&0.8402&0.8720&0.8558\\ \hline
Bi-LSTM CRF&0.7890&0.8066&0.7977&0.8068&\textbf{0.8839}&0.8436\\ \hline
Bi-LSTM CRF-pair&0.8073&\textbf{0.8266}&0.8169&0.8245&0.8527&0.8384\\ \hline
Approximate Skip-Chain CRF&\textbf{0.8364}&0.8062&\textbf{0.8210}&\textbf{0.8614}&0.8651&\textbf{0.8632}\\ \hline
\end{tabular}
\caption{\label{results-table} Cross validated micro-average of Precision, Recall and F-score for all medical tags}
\end{table*}

\noindent{\bf Approximate CRF message passing inference:}
Lin et. al. \shortcite{lin2015deeply} directly estimate the factor to variable message using a Neural Network that uses input image features. Their underlying reasoning is that the factor-to-variable  message from factor $F$ to label variable $y_t$ for any iteration of loopy BP can be approximated as a function of all the input variables and previous messages that are a part of that factor. They only model one iteration of loopy BP, and empirically show that it is leads to an appreciable increase in performance. This allows them to model the messages as a function of only the input variables, since the messages for the first iteration of message passing are computed using the potential functions alone.

We follow a similar approach for calculation of variable marginals in our skip chain model. However, instead of estimating individual factor-to-variable messages, we exploit the sequence structure in our problem and estimate groups of factor-to-variable messages. For any label node $y_t$, the first group, contains factors that involve nodes which occur before $y_t$ in the sentence (from left). The second group of factor-to-variable messages corresponds to factors involving nodes occurring later in the sentence.   We  use recurrent computational units like LSTM to estimate the sum of log factor-to-variable messages within a group. Essentially, we use bidirectional recurrent computation to estimate all the incoming factors from left and right separately.  

To formulate this, let us assume for now that we are using skip edges to connect the current node $t$ to $m$ preceding and $m$ following nodes. Each edge, skip or otherwise, is denoted by a factor which contains the binary potential of the edge and the unary potential of the connected node. As mentioned earlier, we will divide the factors associated with node $t$ into two sets, $F_{L}(t)$ and $F_{R}(t)$. Here $F_{L}(t)$ , contains all factors formed between the variables from the group $\{y_{t-m} , ... , y_{t-1}\}$ and $y_t$. So we can formulate the combined message from factors in $F_{L}(t)$ as 
\begin{equation}
\beta_{L}(y_t)=[ \sum_{F \in F_{L}(t)} \beta_{F \rightarrow t}(y_t)]
\end{equation}
The combined messages from factors in $F_{R}(t)$ which contains variables from $y_{t+1}$ to $y_{t+m}$ can be formulated as :
\begin{equation}
\beta_{R}(y_t)=[ \sum_{F \in F_{R}(t)} \beta_{F \rightarrow t}(y_t)]
\end{equation} 
We also need the unary potential of the label variable $t$ to compose its marginal. The unary potentials of each variable from $\{y_{t-m} , ... , y_{t-1}\}$ and $\{y_{t+1}, ... ,y_{t+m}\}$ should already be included in their respective factors. The log of the unnormalized marginal $\bar{P}(y_t|\textbf{x})$ for the variable $y_t$, can therefore be calculated by  
\begin{equation}
\begin{split}
\log \bar{P}(y_t|\textbf{x}) =\beta_{R}(y_t) +\beta_{L}(y_t) + \phi(y_t)
\end{split}
\end{equation}


Similar to Lin et. al. \shortcite{lin2015deeply}, in the interest of limited network complexity, we use only one message passing iteration. In our setup, this means that a variable-to-factor message from a neighboring variable $y_i$ to the current variable $y_t$ contains only the unary potentials of $y_i$ and binary potential between $y_i$ , $y_t$. As a consequence of this, we can see that $\beta_{L}(y_t)$ can be written as : 
\begin{equation}
\begin{split}
\beta_{L}(y_t) = \sum_{i=t-1}^{t-m} \log \sum_{y_{i}} [\exp \psi(y_{t},y_{i})  + \phi(y_i) ]
\end{split}
\end{equation}
Similarly, we can formulate a function for $\beta_{R}(y_t)$ in a similar way :
\begin{equation}
\begin{split}
\beta_{R}(y_t) = \sum_{i=t+1}^{t+m} \log \sum_{y_{i}} [\exp \psi(y_{t},y_{i})  + \phi(y_i) ]
\end{split}
\end{equation}

\noindent{\bf Modeling the messages using RNN:}
As mentioned previously in equation 8, we only need to estimate $\beta_{+}(y_t)$, $\beta_{+}(y_t)$ and $\phi(\textbf{x})_{t}$ to calculate the marginal of variable $y_t$. We can use $\phi_{nn}(\textbf{x})_{t}$ framework introduced in section 3.2 to estimate the unary potential for $y_t$. We use different directions of a bidirectional LSTM to estimate $\beta_{R}(t)$ and $\beta_{L}(t)$. This eliminates the need to explicitly model and learn pairwise potentials for variables that are not immediate neighbors. 

The input to this layer at position $t$ is  $[\phi_{nn}(y_t) ; \psi_{nn}(y_{t},y_{t+1})]$ (composed of potential functions described in section 3.3). This can be viewed as an LSTM layer aggregating beliefs about $y_t$ from the unary and binary potentials of $[y]_{1}^{t-1}$ to approximate the sum of messages from left side $\beta_{L}(y_t)$.  Similarly, $\beta_{R}(y_t)$ can be approximated from the LSTM aggregating information from the opposite direction. Formally, $\beta_{L}(y_t)$ is approximated as a function of neural network based unary and binary potentials as follows:
\begin{equation}
\beta_{L}(y_t) \approx \textit{f } ([\phi_{nn}(y_i) ; \psi_{nn}(y_{i},y_{i+1})]_{1}^{t-1} )
\end{equation}

Using LSTM as a choice for recurrent computation here is advantageous, because LSTMs are able to learn long term dependencies. In our framework, this allows them to learn to prioritize more relevant potential functions from the sequence $[[\phi_{nn}(y_i) ; \psi_{nn}(y_{i},y_{i+1})]_{1}^{t-1}$. Another advantage of this method is that we can approximate skip edges between all preceding and following nodes, instead of modeling just \textit{m} surrounding ones.  This is because LSTM states are maintained throughout the sentence.

The partition function  for $y_t$  can be easily obtained by using logsumexp over all label entries of the unnormalized log marginal shown in equation 8 as follows:
\begin{equation}
Z_t=\sum_{y_{t}} \exp [\beta_{R}(y_t) +\beta_{L}(y_t) + \phi(y_t)]
\end{equation}
Here the partition function $Z$ is a different for different positions of $t$. Due to our approximations, it is not guaranteed that the partition function calculated from different marginals of the same sentence are equal.  The normalized marginal can be now calculated by normalizing $\log \bar{P}(y_{t}|\textbf{x})$ in equation 8 using $Z_t$.

\begin{equation}
\begin{split}
\mathcal{L}(\textbf{x},\hat{\textbf{y}};\theta)=- \sum_t \sum_{y_t}\delta(y_t=\hat{y}_{t}) (\beta_{R}(y_t;\theta)\\ +\beta_{L}(y_t;\theta) + \phi(y_t;\theta)-\log{Z_t(\theta)})
\end{split}
\end{equation}

The model is optimized using cross entropy loss between the true marginal and the predicted marginal. The loss for a sentence $\textbf{x}$ with a ground truth label sequence $\hat{\textbf{y}}$ is provided in equation 13. 

\section{Dataset}
We use an annotated corpus of 1154 English Electronic Health Records from cancer patients. Each note was annotated by two annotators who label medical entities into several categories. These categories can be broadly divided into two groups, Medical Events and Attributes. Medical events include any specific event that causes or might contribute to a change in a patient's medical status. Attributes are phrases that describe certain important properties about the events.

Medical Event categories in this corpus are \textit{Adverse Drug Event (ADE)}, \textit{Drugname} , \textit{Indication} and \textit{Other Sign Symptom and Diseases (Other SSD)}. \textit{ADE}, \textit{Indication} and \textit{Other SSD} are events having a common vocabulary of Sign, Symptoms and Diseases (SSD). They can be differentiated based on the context that they are used in. A certain SSD should be labeled as \textit{ADE} if it is a side effect of a drug. It is an \textit{Indication} if it is an affliction that a doctor is actively treating with a medication. Any other SSD that does not fall into the above two categories ( for e.g. an SSD in patients history) is labeled as \textit{Other SSD}. \textit{Drugname} event labels any medication or procedure that a physician prescribes. 

The attribute categories contain the following properties, \textit{Severity} , \textit{Route}, \textit{Frequency}, \textit{Duration} and \textit{Dosage}. \textit{Severity} is an attribute of the SSD event types , used to label the severity a disease or symptom. \textit{Route}, \textit{Frequency}, \textit{Duration} and \textit{Dosage} are attributes of \textit{Drugname}. They are used to label the medication method, frequency of dosage, duration of dosage, and the dosage quantity respectively. The annotation statistics of the corpus are provided in the Table 1. 

\section{Experiments}
\begin{figure*}
\begin{center}
   \includegraphics[width=\textwidth]{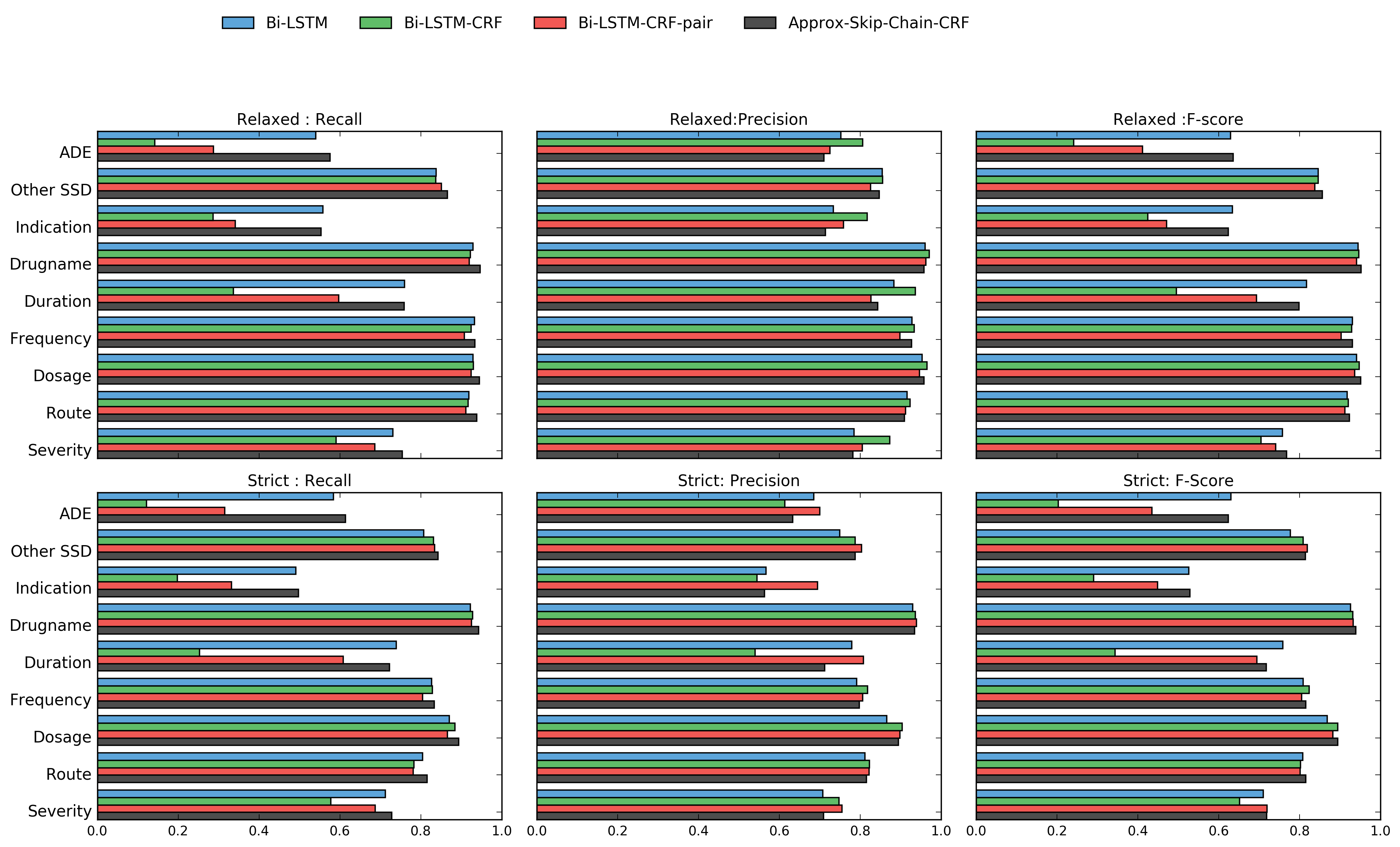}
\end{center}
   \caption{Plots of Recall, Precision and F-score for all four methods. The metrics with prefix \textit{Strict} are using phrase based evaluation. \textit{Relaxed} metrics use word based evaluation.}
\label{fig:loss}
\end{figure*}

Each document is split into separate sentences and the sentences are tokenized into individual word and special character tokens. The models operate on the tokenized sentences. In order to accelerate the training procedure, all models use batch-wise training using a batch of 64 sentences. In order to do this, we restricted the sentence length to 50 tokens. All sentences longer than 50 tokens were cropped to size, and shorter sentences were pre-padded with masks.

The first layer for all models was a 200 dimensional word embedding layer. In order to improve performance, we initialized embedding layer values in all models with a skip-gram word embedding  \cite{mikolov2013distributed}. The skip-gram embedding was calculated using a combined corpus of PubMed open access articles, English Wikipedia and an unlabeled corpus of around hundred thousand Electronic Health Records. The EHRs used in the annotated corpus are not in this unlabeled EHR corpus.

The bidirectional LSTM layer which outputs $\boldsymbol{\omega(\textbf{x})}$ contains LSTM neurons with a hidden size ranging from 200 to 250. This hidden size is kept variable in order to control for the number of trainable parameters between  different models. This helps ensure that the improved performance in models is only because of the modified model structure, and not an increase in trainable parameters. The hidden size is varied in such a way that the number of trainable parameters are close to 3.55 million parameters. Therefore, the Approx skip chain CRF has 200 hidden layer size, while standard RNN model has 250 hidden layer. Since the $\boldsymbol{\omega}(\textbf{x})$ layer is bidirectional, this effectively means that the RNN model has 500 hidden layer size, while Approx skip chain CRF model has 400 dimensional hidden layer. 

We use dropout \cite{srivastava2014dropout} with a probability of 0.50 for all models. We also use batch norm \cite{ioffe2015batch} between layers wherever possible in order to accelerate training. All models are trained in an end-to-end fashion using Adagrad \cite{duchi2011adaptive} with momentum. We use Begin Inside Outside (BIO) label modifiers for models that use CRF based objective.

We use ten-fold cross validation for our results. The documents are divided into training and test documents. From each training set fold, 20\% of the sentences form the validation set which is used for model evaluation during training and for early stopping. 

We report the word based and exact phrase match based micro-averaged recall, precision and F-score. Exact phrase match based evaluation is calculated on a per phrase basis, and considers a phrase as positively labeled only if the phrase exactly matches the true boundary and label of the reference phrase. Word based evaluation metric is calculated on labels of individual words. A word's predicted label is considered as correct if it matches the reference label, irrespective of whether the remaining words in its phrase are labeled correctly. Word based evaluation is a more relaxed metric than phrase based evaluation.

\section{Results}
The micro-averaged Precision, Recall and F-score for all four models are shown in Table 2. We report both strict (exact match) and relaxed (word based) evaluation results. As shown in Table 2, the best performance is Skip-Chain CRF (0.8210 for strict and 0.8632 for relaxed evaluation). Models using exact CRF inference improve the precision of strict evaluation by 2 to 5 percentage points. Bi-LSTM CRF-pair achieved the highest precision for exact-match. However, the recall (both strict and relaxed) for exact CRF models is less than Bi-LSTM. This reduction in recall is much less in the Bi-LSTM-pair model. In relaxed evaluation, only the Skip Chain model has a better F-score than the baseline LSTM. Overall, Bi-LSTM-CRF-pair and Approx-Skip-Chain models lead to performance improvements. However, the standard Bi-LSTM-CRF model does not provide an appreciable increase over the baseline. Figure 1 shows the breakdown of performance for each method with respect to individual medical entity labels. We use pairwise t-test on strict evaluation F-score for each fold in cross validation, to calculate the statistical significance of our scores. The improvement in F-score for Bi-LSTM-CRF-pair and Approx-Skip Chain as compared to baseline is statistically significant ($p < 0.01$). The difference in Bi-LSTM-CRF and baseline, does not appear to be statistically significant ($p > 0.05$).



\section{Discussion}
Overall, Bi-LSTM-CRF-pair and Approx-Skip-Chain CRF models achieved better F-scores than Bi-LSTM and Bi-LSTM-CRF in both strict and relaxed evaluations. The results of strict evaluation, as shown in Figure 1, are our main focus of discussion due to their importance in the clinical domain. As expected, two exact inference-based CRF models (Bi-LSTM-CRF and Bi-LSTM-CRF-pair) show the highest precision for all labels. Approx-Skip-Chain CRF's precision is lower(due to approximate inference) but it still mostly outperforms Bi-LSTM. The recall for Skip Chain CRF is almost equal or better than all other models due to its robustness in modeling dependencies between distant labels. The variations in recall contribute to the major differences in F-scores. These variations can be due to several factors including the rarity of that label in the dataset, the complexity of phrases of a particular label, etc. 

We believe, exact CRF models described here require more training samples than the baseline Bi-LSTM to achieve a comparable recall for labels that are complex or ``difficult to detect''. For example, as shown in table 1, we can divide the labels into frequent ( \textit{Other SSD}, \textit{Indication}, \textit{Severity}, \textit{Drugname}, \textit{Dosage}, and \textit{Frequency}) and rare or sparse (\textit{Duration}, \textit{ADE}, \textit{Route}). We can make a broad generalization, that exact CRF models (especially Bi-LSTM-CRF) have somewhat lower recall for rare labels. This is true for most labels except for \textit{Route}, \textit{Indication}, and \textit{Severity}. The CRF models have very close recall (0.780,0.782) to the baseline Bi-LSTM (0.803) for \textit{Route} even though its number of incidences are lower (2,862 incidences) than \textit{Indication} (3,724 incidences) and \textit{Severity} (3,628 incidences, Table 1), both of which have lower recall even though their incidences are much higher. 


Complexity of each label can explain the aforementioned phenomenon. \textit{Route} for instance, frequently contains unique phrases such as ``by mouth" or ``p.o.," and is therefore easier to detect. In contrast, \textit{Indication} is ambiguous. Its vocabulary is close to two other labels: \textit{ADE} (1,807 incidences) and the most populous \textit{Other SSD} (40,984 incidences). As a consequence, it is harder to separate the three labels. Models need to learn cues from surrounding context, which is more difficult and requires more samples. This is why the recall for \textit{Indication} is lower for CRF models, even though its number of incidences is higher than \textit{Route}. To further support our explanation, our results show that the exact CRF models mislabeled around 40\% of \textit{Indication} words as \textit{Other SSD}, as opposed to just 20 \% in case of the baseline. The label \textit{Severity} is a similar case. It contains non-label-specific phrases such as ``not terribly", ``very rare" and ``small area," which may explain why almost 35\% of \textit{Severity} words are mislabeled as \textit{Outside} by the bi-LSTM-CRF as opposed to around 20\% by the baseline.

It is worthwhile to note that among exact CRF models, the recall for Bi-LSTM-CRF-pair is much better than Bi-LSTM-CRF even for sparse labels. This validates our initial hypothesis that Neural Net based pairwise modeling may lead to better detection of rare labels. 

\section{Conclusion}
We have shown that modeling pairwise potentials and using an approximate version of Skip-chain inference increase the performance of the Bi-LSTM-CRF model.These results suggest that the structured prediction models are good directions for improving the exact phrase extraction for medical entities.

\section*{Acknowledgments}
We thank the UMassMed annotation team: Elaine Freund, Wiesong Liu, Steve Belknap, Nadya Frid, Alex Granillo, Heather Keating, and Victoria Wang for creating the gold standard evaluation set used in this work.  We also thank  the anonymous reviewers for their comments and suggestions.

This work was supported in part by the grant HL125089 from the National Institutes of Health (NIH). We also acknowledge the support from the United States Department of Veterans Affairs (VA) through Award 1I01HX001457. This work was also supported in part by the Center for Intelligent Information Retrieval. The contents of this paper do not represent the views of CIIR, NIH, VA or the United States Government.  

\bibliography{rnn-crf}
\bibliographystyle{emnlp2016}

\end{document}